\documentclass[letterpaper, 10 pt, conference]{ieeeconf}  % Comment this line out if you need a4paper

\IEEEoverridecommandlockouts                              % This command is only needed if 
\usepackage{graphicx}
\graphicspath{ {./pics/} }
\usepackage[caption=false]{subfig}
\usepackage{amsmath}

\title{Efficient Latent Representations using Multiple Tasks\\ for Autonomous Driving}

\author{Eshagh Kargar$^{1}$ and Ville Kyrki$^{1}$% <-this % stops a space
\thanks{$^{1}$Eshagh Kargar and Ville Kyrki are with School of Electrical Engineering, Aalto University, Finland. {firstname.lastname}@aalto.fi}%
}

\begin{document}

\maketitle
\thispagestyle{empty}
\pagestyle{empty}

%%%%%%%%%%%%%%%%%%%%%%%%%%%%%%%%%%%%%%%%%%%%%%%%%%%%%%%%%%%%%%%%%%%%%%%%%%%%%%%%
\begin{abstract}

Driving in the dynamic, multi-agent, and complex urban environment is a difficult task requiring a complex decision policy. The learning of such a policy requires a state representation that can encode the entire environment. Mid-level representations that encode a vehicle's environment as images have become a popular choice, but they are quite high-dimensional, which limits their use in data-scarce cases such as reinforcement learning. In this article, we propose to learn a low dimensional and rich feature representation of the environment by training an encoder-decoder deep neural network to predict multiple application relevant factors such as trajectories of other agents. We demonstrate that the use of the multi-head encoder-decoder neural network results in a more informative representation compared to a single-head encoder-decoder model. In particular, the proposed representation learning approach helps the policy network to learn faster, with increased performance and with less data, compared to existing approaches using a single-head network. 
%We propose a multi-task model to learn task-specific feature representation for a dynamic and multi-agent urban driving environment. Rather than purely using an encoder-decoder structure to reconstruct the input scene, we use multiple heads for some additional tasks such as trajectory planning for the car itself and other agents. Considering these multiple heads will force the network to learn about possible future trajectories for the vehicle itself and other agents around it. We show that using multi-head network and auxiliary tasks helps the policy network to learn faster and perform better, with much lower data, compared to the single-head network. 
\end{abstract}

%%%%%%%%%%%%%%%%%%%%%%%%%%%%%%%%%%%%%%%%%%%%%%%%%%%%%%%%%%%%%%%%%%%%%%%%%%%%%%%%
\section{INTRODUCTION}

% Introduction to autonomous driving.
Driving in unstructured and dynamic urban environments is an arduous task. There are a lot of moving agents such as cars, bicycles, and pedestrians that affect driver behaviors and decisions. To drive a car, a driver, whether a human or artificial, needs to perceive and understand other agents' behaviors, plans, and the interactions in the environment, and plan based on that understanding. The number of the factors makes the state space of this problem very large, and driving safely in this environment is an open challenge for the research community and industry.

% Representation problem, from end-to-end to mid-to-mid
A central challenge related to the high complexity of the traffic environment is how to represent the environment state. In particular, if the decision policies are learned from human imitation or through reinforcement learning, the environment needs to be represented as a state vector of constant dimension, to allow it to be used as input to a policy function represented for example as a neural network. Immediate sensor measurements are used directly as the state in the so called end-to-end methods. However, the very high dimensionality of the sensor data makes its direct use challenging, as vast number of data are needed to constrain the learning problem \cite{muller2018driving}. For that reason, mid-to-mid approaches that encode a vehicle's environment as a single image have recently received increasing attention (see e.g.~\cite{bansal2018chauffeurnet}). However, even the mid-level representations are quite high-dimensional, which limits their use in data scarce cases such as reinforcement learning. 

\begin{figure}[t]
\includegraphics[scale=0.4]{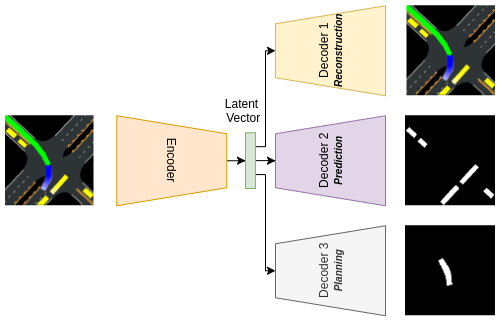}
\centering
\caption{Multi-head network architecture.}
\label{fig:multi_head}
\end{figure}

% Approach
To alleviate this, we propose a new approach to learn low-dimensional and rich  representations. In particular, we combine the recently proposed idea to learn a low-dimensional latent space of the mid-level image \cite{chen2019model} with prediction of multiple auxiliary application relevant factors such as trajectories. The single latent representation is extracted from the mid-level image, but the latent representation is enforced to predict the trajectories of both the ego vehicle and other vehicles in addition to the input image, using a multi-head network structure as depicted in Fig.~\ref{fig:multi_head}. All heads of the network represent the information as images which allows their easy interpretation. However, it is important to note that only the encoder part of the network is used to extract the latent state space vector and the decoders are not used in the policy training.

% Benefits
Experiments demonstrate that the auxiliary tasks allow the network to learn more representative information from the scene. Therefore, a policy network can be trained faster and it performs better, even with much lower data, compared to a representation based on a single-head network that is trained for the optimal reconstruction of the input.

% Contributions
The primary contributions of this work are:
\begin{itemize}
    \item a multi-task network with auxiliary heads to improve the quality of low-dimensional representations for mid-to-mid autonomous driving approaches,%to do planning for the car and prediction for other agents' future movements in order to consider agents movements learn more representative latent space,
    \item an experimental study of policy imitation performance, showing that by using auxiliary tasks, the policy can be (i) trained faster, (ii) it performs better, and (iii) the learning of the policy requires less data.
\end{itemize}

Our paper is organized as follows: in Section II we review related works; in Section III we explain the input representation and model architectures; in Section IV we talk about the dataset, implementation, and architecture details of the networks. Then, we discuss the results in Section V and finally the conclusion in Section VI. 

\section{RELATED WORK}
% Explain about methods that use camera images for imitation learning.
With recent advances in deep learning, many works have addressed the decision making problem in autonomous driving systems by attempting to map sensor observations directly to decisions or control commands.
Multiple authors \cite{bojarski2016end}, \cite{xu2017end} have proposed methods to map camera images directly to control commands using neural networks. 
Muller et al.~\cite{muller2018driving} proposed to first apply semantic segmentation and then use the segmented image as the input to the driving policy, in order to provide robustness across imaging conditions.
Camera images can also be used to estimate affordances such as distance to the preceding car which can be used to program a controller to control the car on a highway \cite{chen2015deepdriving}.
Camera images can also be integrated with some other information such as navigational commands to learn a mapping to control commands \cite{wang2018end}, \cite{codevilla2018end}.
Some other works tried to do several tasks simultaneously, such as Luo et al.~in \cite{luo2018fast} which used an end-to-end neural network to map 3D sensor data to 3D detection, tracking, and motion forecasting tasks simultaneously.
Hawke et al.~also used camera images to output segmentation, monocular depth, and optical flow and then used the mid-level feature-map and route command to train a policy to output control commands \cite{hawke2019urban}.

% explain about mid-to-mid models and how they extract features.
Recently, mid-to-mid methods have gained increasing popularity. 
Bansal et al.~proposed a mid-to-mid method to do imitation learning and data augmentation to learn the driving policy \cite{bansal2018chauffeurnet}.
To address the problem of high-dimensional state space, Chen et al.~\cite{chen2019model} trained a variational auto-encoder (VAE) to reconstruct the input bird-eye view image, with rendered mid-level information on it, and used the latent vector as input to their reinforcement learning agent to output control commands. 
Chen et al.~\cite{chen2019deep} used mid-level information and trained a network to output a trajectory. They also proposed safety and tracking controllers to get the planned trajectory and output a safe control command.
There are some other works for motion forecasting such as \cite{cui2019multimodal, cui2019deep, chou2019predicting, phan2019covernet, chai2019multipath} that used mid-level information and rendered them on a bird-view image as input to their algorithms. 

Our work fits in the mid-to-mid imitation learning, motion prediction, and multi-task learning literature. We extend the representation learning capability of latent space by using auxiliary tasks such as planning for the ego car and other agents, and show that by considering these tasks, the policy can be learned very fast with much lower data.

%%%%%%%%%%%%%%%%%%%%%%%%%%%%%%%%%%%%%%%%%%%%%%%%%%%%%%%%%%%%%%%%%%%%%%%%%%%%%%%%
\section{Method}

Learning a driving policy for dense urban environments directly from raw sensor data is a difficult task and needs a lot of data. To reduce the complexity of the scene, we use bird-view image with rendered information on it. In our simulation experiments, the perception and routing information is provided by the CARLA simulator \cite{dosovitskiy2017carla}. Then we render this information on a bird-view image $x$ similar to e.g.~\cite{chen2019model}. We then use an encoder $f_\theta(z|x)$, which is trained using single-head or multi-head network, to encode this bird-view image into a low-dimensional latent vector $z$. The driving policy $u=\pi(z)$ is then trained to map this latent vector into control commands $u$. 
% The general structure of the system is shown in Fig. \ref{fig:system}

% \begin{figure}[t]
% \includegraphics[scale=0.4]{pics/system.png}
% \centering
% \caption{System architecture}
% \label{fig:system}
% \end{figure}

\subsection{Input Representation}
% explain about the input data and which information is used and rendered on the image.
The perception and routing information that we use to generate the bird-view image and feed it into the encoder network are as follows: 
\begin{itemize}
    \item \textbf{Map}: This information contains different data from HD-Maps, such as road boundary, lane lines, curbs, etc.
    \item \textbf{Route}: This information is rendered on the image with green or red colors. The color represents the traffic lights state. Red is for the red traffic light, and the green is for green one. This will be useful in some scenarios, such as intersections.
    \item \textbf{Traffic lights}: This information is rendered on the image with green or red colors on the planned route, as we described earlier.
    \item \textbf{Current and past ego agent poses}: We render current and past poses of the ego vehicle on the image with blue color.
    \item \textbf{Current and past poses of dynamic objects in the environment}: We render current and past poses of other agents on the image with yellow color.
\end{itemize}

The field of view is considered to be 40m*40m, and the ego car is positioned at (20m, 15m). We also rendered ten past poses for the ego vehicle and other agents on the bird-view image with lower color intensity. By using this type of information representation, it is easy to consider agents' sizes, their headings, different distances in the scene, road curves, drivable areas, etc. The example of this input representation can be seen as the input in Fig. \ref{fig:multi_head}.

\subsection{Model architecture}

To reduce the dimensionality of the input to the policy and learn a driving policy and prevent over-fitting in some methods such as reinforcement learning algorithms, we use an encoder-decoder structure. The encoder part maps the high dimensional bird-view image into a low dimensional latent vector. 

We propose a multi-head decoder that consists of several networks which decode the low-dimensional latent vector back into the bird-view image and some other images. 
As a comparison baseline, we use a single-head auto-encoder network to compare with the proposed multi-head network, similar to \cite{chen2019model}. A simple policy network is used for evaluation of the learned latent space for the encoder-decoder models.
In the next sections, we describe the structure of single-head, multi-head, and policy networks.

\subsubsection{Single-Head Encoder-Decoder}

\begin{figure}[t]
\includegraphics[scale=0.4]{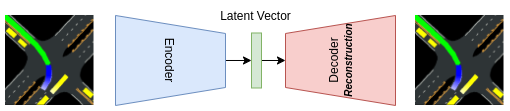}
\centering
\caption{Single-head network architecture}
\label{fig:single_head}
\end{figure}

This network has one encoder and one decoder. The encoder network \(f_\theta (z|x) \) which gets the input bird-view image \(x\) and encodes it into a low dimensional latent vector \(z\) and the decoder network \( g_\phi (\hat{x} | z) \) that gets the encoded latent vector \(z\) and outputs the reconstructed bird-view image \(\hat{x}\). Here \(\theta\) and \(\phi\) denote the parameters of the encoder and decoder networks, respectively. The network architecture is given in Fig.\ref{fig:single_head}

To train this encoder-decoder structure and obtain the \(\theta\) and \(\phi\) parameters, we used the following objective function to minimize the error between the input bird-view image and the reconstructed image as follows:

\[L_{single\_{head}} = -E_{f_\theta (z|x)} log (g_\phi (\hat{x} | z))) \]

%% put an image of the reconstructed and gt images

\subsubsection{Multi-Head Encoder-Decoder}

The other encoder-decoder neural network which we used to encode the bird-view image into a low dimensional latent vector has one encoder and three decoders as follows:

\textbf{Reconstruction Head} is similar to the decoder network of the single-head model for reconstructing the input bird-view image.

\textbf{Prediction Head} is a network to make the motion forecasting for other agents in the scene. It gets the latent vector and tries to generate a binary image of the \(1s \) future trajectory of other agents.

\textbf{Planning Head} is a network which tries to do the planning for the ego vehicle and generate a binary image of future trajectory for \(1s \). 

The encoder network \(p_\psi (v|x) \) gets the input bird-view image \(x\) and encodes it into the latent vector \(v\). Then we considered several decoder networks to get the latent vector \(v\) and reconstruct the bird-view image using \( q_{1_{\eta_1}} (\hat{x}|v) \), predict other agents' future motions using \( q_{2_{\eta_2}} (pred|v) \), and do planning for the ego vehicle using \( q_{3_{\eta_3}} (plan|v) \), where \(\hat{x} \) is the reconstructed bird-view image, and \(pred\) and \(plan\) are two binary images with the future predicted motion for other agents in the scene and the planned trajectory for the ego vehicle, respectively. Here \(\eta_1, \eta_2, \eta_3 \) are the parameters for the reconstruction, motion prediction, and planning heads of three decoder networks. 
The network architecture for the multi-head model is shown in Fig.\ref{fig:multi_head}.

The objective function for this multi-head network which needs to be minimized is a summation of three loss functions between the encoder and three decoder heads, as follows:
\[ L_{multi\_head} = -E_{p_{\psi} (v|x)} log (q_{1_{\eta_1}} (\hat{x}|v)) - \]
\[ E_{p_\psi (v|x)} log (q_{2_{\eta_2}} (pred|v)) - \]
\[ E_{p_\psi (v|x)} log (q_{3_{\eta_3}} (plan|v)) \]

\subsubsection{Policy Network}

To evaluate the performance of each encoder-decoder model, we use a simple two-heads fully connected policy network to predict steering angle and acceleration.
The steering angle value is a continuous number, and we use a smooth L1 loss function \cite{girshick2015fast} as its objective function. The acceleration values are only three discrete values. So we consider it as a classification problem and use a cross-entropy loss function for it. The loss function for policy network is as follows:
\[ L_{policy} = L_{1_{smooth}}(steer, steer_{gt}) +\]
\[CrossEntropy(acc, acc_{gt})  \]
% where:
% \[ L_1_{smooth}(x, y) = \frac{1}{n} \sum_{i} z_i \]
%     \[z_i = \begin{cases}
%     0.5(x_i - y_i)^2, & \text{if}\ |x_i - y_i| < 1 \\
%     |x_i - y_i| - 0.5 , & \text{otherwise}
%     \end{cases}
%     \]
% \[ CrossEntropy (x, y) = -E_{x,y~\hat{p}_{data}} log(p_{model}(y|x) \]
where \(steer\) and \(acc\) are the outputs of the policy network for steering angle and acceleration, respectively, and \(steer_{gt}\) and \(acc_{gt}\) are ground truth values for steering angle and acceleration.

%%%%%%%%%%%%%%%%%%%%%%%%%%%%%%%%%%%%%%%%%%%%%%%%%%%%%%%%%%%%%%%%%%%%%%%%%%%%%%%%
\section{Experiments}

\subsection{Simulation Environment and Data Collection}
% Explain about data and simulator
To collect the dataset to train these two models, we used the CARLA simulator, which is an open-source simulator for autonomous driving, and spawned 100 vehicles randomly in the Town Three and used CARLA autopilot to collect a driving dataset with 500k frames. In each time step, we render different information on the bird-view image. By recording the pose of all agents in each time-step during driving, we have the required information to create the dataset without any manual labeling. For the input bird-view image, we render \(1s \) of the pose history of the ego vehicle and other agents together with other information such as map information, planned route, and traffic light state on the input image. To create the ground truth data for prediction and planning heads, we can only use the pose of the ego car and other agents in the next time-steps and transform them into the current ego vehicle's frame and render them on a binary image. So no manual labeling is required, and it can be done in a self-supervised manner.

For policy training, we noticed that the dataset, especially for steering angle, which is zero most of the time, is highly imbalanced. So we created a new dataset and balanced it using sub-sampling. The size of the policy dataset is about 80k data samples.

The size of all images, the bird-view input image and prediction and planning binary images, is 256 x 256, which is resized to 64 x 64 to feed them into the models.

\subsection{Network Architectures}

To have a fair comparison between the two models, we consider the same structure for the encoder network in both models. Also, the general decoder network for all tasks of reconstruction, prediction, and planning is the same, except at the final layer of the decoder, which will generate an RGB image or binary image.

\subsubsection{Single-Head Encode-Decoder}
This network has one encoder network followed by one decoder network and tries to reconstruct the input bird-view image.

The encoder network has three Conv-layers of 4 x 4 kernel size, stride 2, with 32, 64, and 128 channels. Each Conv-layer is followed by a BatchNorm layer and ReLU activation function. Then the feature-map with the size of 128 x 6 x 6 is fully connected to the latent space with the size of 64.

The decoder network, the reconstruction head, has one fully connected layer followed by BatchNorm-layer and ReLU activation function and a reshape function to generate the feature-map with the size of 128 x 6 x 6 , and then three ConvTranspose-layers with stride 2 and kernel size 4 x 4. The first and the second ConvTranspose-layers have 64 and 32 channels and are followed by BatchNorm-layer and ReLU activation function. The last ConvTranspose-layer has three channels and is followed by a Sigmoid activation function.

\begin{figure}[h!]
\includegraphics[width=\columnwidth]{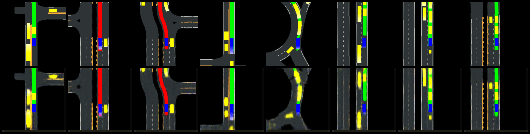}
\centering
\caption{Single-head network reconstruction results. The first row shows the original input images and the second row shows the output results.}
\label{fig:single_head_rec}
\end{figure}

The single-head encoder-decoder network is trained from scratch using Adam \cite{kingma2014adam} optimizer with learning rate \( 5^{-3} \) and batch size 2048 for 200 epochs.
Some examples of the trained single-head encoder-decoder network are given in Fig. \ref{fig:single_head_rec}.

\subsubsection{Multi-Head Encoder-Decoder}

The multi-head encoder-decoder network has one encoder and three decoder heads: reconstruction head, planning head, and prediction head.

The network architecture of the encoder part is similar to the encoder network of the single-head model, and the decoder heads have a similar architecture as the decoder network of the single-head model. The only difference is the number of channels of the last ConvTranspose-layers which is one for prediction and planning heads and three for the reconstruction head.  

\begin{figure}[h!]
\centering
\subfloat[Reconstruction results]{%
  \includegraphics[width=\columnwidth]{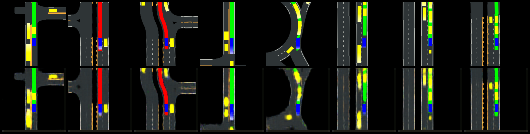}%
}

\subfloat[Prediction results]{%
  \includegraphics[width=\columnwidth]{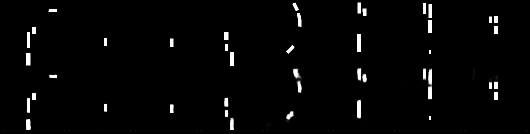}%
}

\subfloat[Planning results]{%
  \includegraphics[width=\columnwidth]{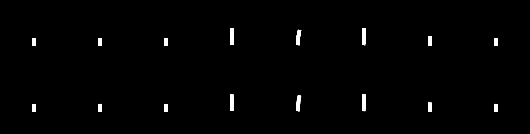}%
}

\caption{Multi-head network output results. The first row in each figure shows the original input image and the second row shows the outputs of the network.}
\label{fig:multi_results}
\end{figure}

The training hyper-parameters of the multi-head network are the same as the single-head network. Some results of the trained model for reconstruction, prediction, and planning heads are shown in Fig. \ref{fig:multi_results}.

\subsection{Policy Network Architecture}
To evaluate the performance and representation learning capability of two models, we also trained a simple policy network to get the normalized latent vector as input and output steering angle and acceleration. 

The policy network has two heads for steering angle and acceleration prediction. The steering angle head has three fully connected layers with 256, 64, and one neuron. The first two layers are followed by ReLU activation function and Dropout layer with a dropout probability of 0.5, and the last layer has only a linear activation function. The output of this head is a continuous value that can be clipped between some specific values to be applied to the car. Here we crop it between -0.25 to 0.25.

\begin{figure}[h!]
\includegraphics[scale=0.4]{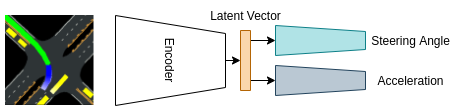}
\centering
\caption{Encoder and policy networks for steering angle and acceleration prediction. The encoder can be trained using each one of single-head or multi-head models.}
\label{fig:policy}
\end{figure}

The acceleration head has three fully connected layers with 128, 64, and three neurons. The first two layers are followed by ReLU activation function and Dropout layer with a dropout probability of 0.5, and the last layer has only a Softmax activation function to generate a probability distribution for three classes, which we considered for acceleration: -1, 0.5, 1. 
The final structure with the policy and the encoder is shown in Fig. \ref{fig:policy}.

The policy network is trained from scratch using Adam \cite{kingma2014adam} optimizer with learning rate \( 5^{-4} \) and batch size 2048 for 100 epochs.

% \begin{figure}[h!]
% \centering
% \subfloat[Steering angle loss]{%
%   \includegraphics[width=0.8\columnwidth]{pics/steer_train_compare.png}%
% }

% \subfloat[Acceleration accuracy]{%
%   \includegraphics[width=0.8\columnwidth]{pics/acc_train_compare.png}%
% }
% \caption{Train performance of trained policies using the encoder form trained encoder-decoder networks with different heads.}
% \label{fig:train_compare_heads}
% \end{figure}

\begin{figure}[h!]
\centering
\subfloat[Steering angle loss]{%
  \includegraphics[width=\columnwidth]{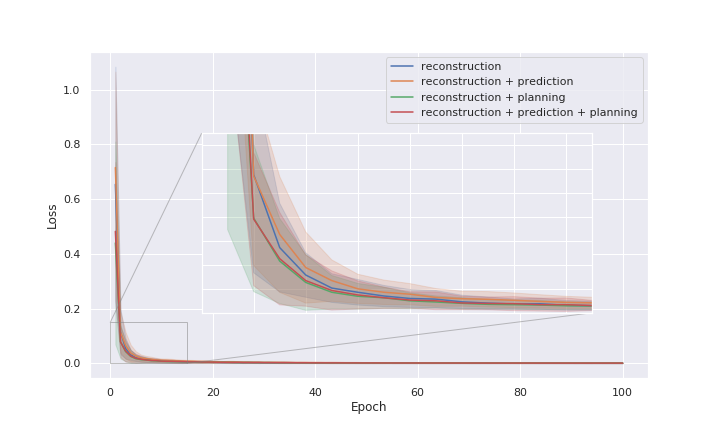}%
}

\subfloat[Acceleration accuracy]{%
  \includegraphics[width=\columnwidth]{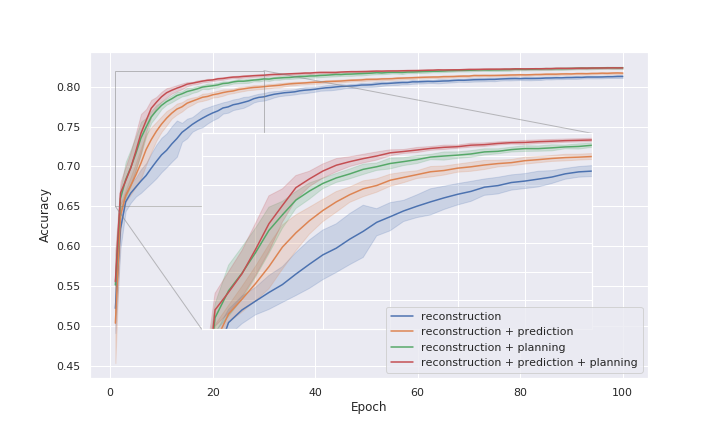}%
}
\caption{Test performance of trained policies using the encoder form trained encoder-decoder networks with different heads. 
The shaded areas in the plot show the variance.}
\label{fig:test_compare_heads}
\end{figure}

\section{Results}
% compare the proposed method with baseline method.
To evaluate the latent vector learned by the proposed multi-head network, we consider different cases such as evaluating the effect of different heads and the size of the dataset and train the policy network in each case. 

All the models are implemented in PyTorch \cite{paszke2019pytorch} and trained on two NVIDIA RTX 2080Ti GPUs.
In all of the training procedures, for encoder-decoder and policy networks, we used 80\% of the dataset for training and 20\% for test. For evaluating the model with different dataset sizes, we selected a subset of data from 80\% training data randomly. 

In the following sections, we explain about different cases and discuss the results.

\subsection{Effect of Different Heads}

In this case, to see the effect of each head in the learned latent vector, we train a single-head network with only the reconstruction head, two-heads network with reconstruction and prediction heads, two-heads network with reconstruction and planning heads, and full model with all three reconstruction, prediction, and planning heads. All the encoder-decoder networks are trained using the full dataset with 500k data samples. Then we use the encoder of each one of these trained models as feature extractor and train the policy network. All policies are trained on another dataset with 80k data samples.

Fig. \ref{fig:test_compare_heads} shows the results of steering angle and acceleration prediction for different cases and for 15 times training from scratch, 100 epochs each, with different initialization weights. As can be seen, the performance of the steering head is good for all cases, but for the acceleration head, performance improvement is clear by adding heads one by one. The full model with three heads has the best performance with faster convergence and reaches the final accuracy of single-head model in about 20 epochs, one-fifth of training time of the single-head model.
The encoder-decoder network with reconstruction and planning heads is the second best model. By comparing two two-heads models, reconstruction plus prediction and reconstruction plus planning, it can be seen that the effect of adding the planning head to the single-head model is more than the prediction head. From the figures, it is also clear that the variance of performance for the single-head model in transient time is more than others.

\begin{figure}[h!]
\centering
\subfloat[Steering angle loss]{%
  \includegraphics[width=0.8\columnwidth]{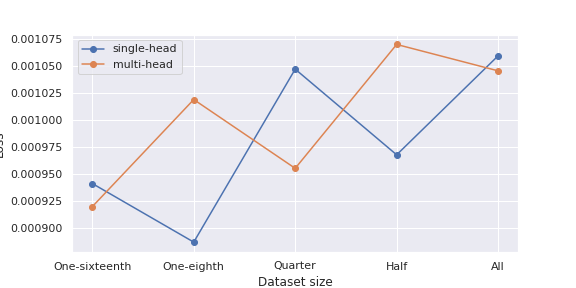}%
}

\subfloat[Acceleration accuracy]{%
  \includegraphics[width=0.8\columnwidth]{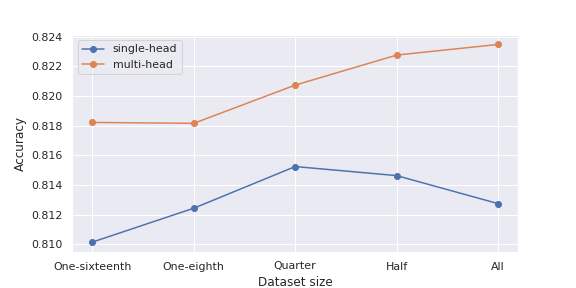}%
}

\caption{Test performance of trained policies using the encoder form trained encoder-decoder networks with different dataset sizes.}
\label{fig:datasize_compare}
\end{figure}

\subsection{Effect of Dataset Size}

\begin{figure}[h!]
\centering
% \subfloat[Acceleration accuracy - Train]{%
%   \includegraphics[width=0.8\columnwidth]{pics/acc_train_compare_datasize.png}%
% }

% \subfloat[Acceleration accuracy]{%
  \includegraphics[width=\columnwidth]{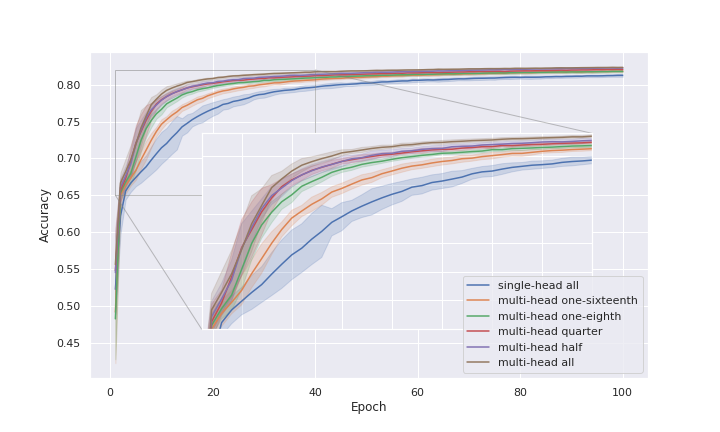}%
% }
\caption{Acceleration accuracy of trained policies using the encoder form trained encoder-decoder networks with different dataset sizes. The shaded areas in the plot show the variance.}
\label{fig:datasize_convergespeed}
\end{figure}

In this case, we train two models, single-head and full multi-head models, using the full dataset, which is 500k frames, 50\% of the dataset, a 25\% of the dataset, 12.5\% of the dataset, and 6.25\% of the dataset, and then use the trained encoder of each model as feature extractor for training the policy. All policies are trained for 15 times from scratch , 100 epochs each, on another dataset with 80k data samples. Fig. \ref{fig:datasize_compare} shows the mean performance of the test phase for different dataset sizes for the last five epochs. As can be seen, the steering angle loss in all cases are the same and have only very small difference. For acceleration prediction, the performance of the multi-head network trained on all dataset sizes is better than the single-head model, and has better accuracy even with 6.25\% of the data.
In addition, the convergence speed of the multi-head network trained on all dataset sizes, for acceleration prediction, is compared to the single-head model trained on the full dataset and is showed in Fig. \ref{fig:datasize_convergespeed}. As can be seen in this figure, multi-head models perform better and converge faster compared to the single-head model. The multi-head model can reach the single-head model's performance in 20 epochs, one-fifth of training time of the single-head model, with full dataset, in 40 epochs with a quarter of the dataset, and in 60 epochs using one-sixteenth of the dataset. In general, the multi-heal model, using only 6.25\% of the dataset, converges faster and perform better than single head model trained on the full dataset.

%-----------------------------------------------

%%%%%%%%%%%%%%%%%%%%%%%%%%%%%%%%%%%%%%%%%%%%%%%%%%%%%%%%%%%%%%%%%%%%%%%%%%%%%%%%
\section{CONCLUSIONS}

In this paper, we proposed a multi-head network using auxiliary tasks for latent space representation learning that can be used to train a policy network using imitation or reinforcement learning. We used a bird-view input representation to render the required information from an HD-map and a perception module on an image that was then encoded into a latent representation. Experimental comparison against a baseline showed that the use of auxiliary tasks in representation learning improves policy learning in three aspects: the policy quality is better, the learning converges faster, and the learning of the policy requires fewer data. The proposed approach can be trivially extended to other task-relevant factors if they can be encoded as images. However, the question of how to balance the auxiliary tasks in cases where their representations differ requires further study.

\addtolength{\textheight}{-12cm}   % This command serves to balance the column lengths



%%%%%%%%%%%%%%%%%%%%%%%%%%%%%%%%%%%%%%%%%%%%%%%%%%%%%%%%%%%%%%%%%%%%%%%%%%%%%%%%
% \section*{ACKNOWLEDGMENT}

%%%%%%%%%%%%%%%%%%%%%%%%%%%%%%%%%%%%%%%%%%%%%%%%%%%%%%%%%%%%%%%%%%%%%%%%%%%%%%%%

\bibliographystyle{IEEEtran}
\bibliography{mine}

\end{document}